\documentclass{article}

\usepackage{PRIMEarxiv}

\usepackage[utf8]{inputenc} 
\usepackage[T1]{fontenc}    
\usepackage{hyperref}       
\usepackage{url}            
\usepackage{booktabs}       
\usepackage{amsfonts}       
\usepackage{nicefrac}       
\usepackage{microtype}      
\usepackage{lipsum}
\usepackage{fancyhdr}  
\usepackage{algorithm,algpseudocode}
\usepackage{caption}
\usepackage{graphicx}   
\usepackage{multirow}
\usepackage{commath}

\graphicspath{{media/}}     

\title{A Comparative Study of Calibration Methods for Imbalanced Class Incremental Learning}

\author{Umang Aggarwal$^{1,2}$, Adrian Popescu$^{1}$, Eden Belouadah$^{1}$, Céline Hudelot$^{2}$\\
\small{(1) Université Paris-Saclay, CEA, Département Intelligence Ambiante et Systèmes Interactifs}\\
\small{(2) 
Université Paris-Saclay, CentraleSupélec, Mathématiques et Informatique pour la Complexité et les Systèmes} \\
\small{91191 Gif-sur-Yvette, France}\\
{\tt\small {umang.aggarwal,adrian.popescu}@cea.fr,celine.hudelot@centralesupelec.fr}
}

\begin{document}
\maketitle

Deep learning approaches are successful in a wide range of AI problems and in particular for visual recognition tasks.
However, there are still open problems among which is the capacity to handle streams of visual information and the management of class imbalance in datasets.
Existing research approaches these two problems separately while they co-occur in real world applications.
Here, we study the problem of learning incrementally from imbalanced datasets.
We focus on algorithms which have a constant deep model complexity and use a bounded memory to store exemplars of old classes across incremental states.
Since memory is bounded, old classes are learned with fewer images than new classes and an imbalance due to incremental learning is added to the initial dataset imbalance.
A score prediction bias in favor of new classes appears and we evaluate a comprehensive set of score calibration methods to reduce it.  
Evaluation is carried with three datasets, using two dataset imbalance configurations and three bounded memory sizes. 
Results show that most calibration methods have beneficial effect and that they are most useful for lower bounded memory sizes, which are most interesting in practice. 
As a secondary contribution, we remove the usual distillation component from the loss function of incremental learning algorithms. 
We show that simpler vanilla fine tuning is a stronger backbone for imbalanced incremental learning algorithms.


\section{Introduction} 
\label{sec:intro}
 
Large scale image repositories are highly dynamic, with content being added and/or removed at a fast pace.
However, content analysis is currently done with algorithms built to learn from static information. 
This is notably the case for deep learning models which are trained on fixed datasets.
When model updates are required, the entire training corpus is reused for learning, making the process cumbersome. 
To make image content analysis more dynamic and thus more adapted to dynamic corpora, incremental \cite{DBLP:conf/cvpr/RebuffiKSL17} or lifelong \cite{DBLP:conf/cvpr/AljundiCT17} learning processes need to be implemented. 

In this paper, we focus on the intersection of incremental learning and class imbalanced learning.
Each problem is well studied separately but, to our knowledge, they were not tackled together in the context of deep learning. 
A joint study is needed to cope with dynamic and imbalanced datasets.
The main challenge in incremental learning is due to a restricted or impossible access to old data. 
This restriction can be explained by factors such as: data removal on the Web and in stream data processing \cite{DBLP:conf/esann/HammerHM14}, privacy in the medical domain \cite{DBLP:journals/corr/VenkatesanVPL17}, limited resources in embedded systems \cite{continual_review:2019}.
Recent research in incremental learning use deep learning as backbone and most of them focus on class incremental learning, a setting in which data are completely labeled. 
If no memory of the past is allowed, the deep architecture grows in time to accommodate new classes \cite{DBLP:conf/cvpr/AljundiCT17,DBLP:journals/corr/RusuRDSKKPH16,DBLP:conf/cvpr/WangRH17}.
If a bounded memory is allowed, the architecture is fixed and an adapted fine tuning is applied to learn incrementally \cite{DBLP:conf/eccv/CastroMGSA18,DBLP:journals/corr/abs-1807-02802,DBLP:conf/cvpr/RebuffiKSL17}.

A wide majority of deep learning methods assume that training datasets are balanced or nearly so. 
This is for instance the case of the ImageNet $LSVRC$ effort \cite{DBLP:journals/ijcv/RussakovskyDSKS15}, in which 1000 leaf classes which are well represented in the dataset and rather balanced. The $ILSVRC$ training set used in \cite{DBLP:conf/cvpr/RebuffiKSL17} has a mean of 1231 images per class, with a standard deviation of 70.
However, an analysis of the full set of ImageNet leaf classes \cite{DBLP:conf/cvpr/DengDSLL009} shows that image counts per class are highly variable, with a mean of 592 and a standard deviation of 508. 
The same is true for most existing large scale public datasets, including Open Images \cite{OpenImages2}, Google Landmarks \cite{DBLP:conf/iccv/NohASWH17} or MS-CELEB-1M \cite{DBLP:conf/eccv/GuoZHHG16}.
The datasets built for real-life applications are often imbalanced and the classes of interest are often under-represented.
This is, for instance, the case in the medical domain or in fraud detection. 
Learning from imbalanced datasets which include minority and majority classes leads to a prediction bias towards the majority classes.
This was shown by two surveys of classical machine learning methods devised for imbalanced datasets \cite{DBLP:journals/tkde/HeG09,DBLP:journals/ida/JapkowiczS02}. 
A similar conclusion was recently presented in \cite{DBLP:journals/nn/BudaMM18}, where the authors study the effect of data imbalance on deep learning algorithms.

Here we advocate that class incremental learning with a bounded memory actually boils down to a form of imbalanced learning problem.
New data most often corresponds to majority classes, while old data corresponds to minority classes since images of old classes need to be fit in the bounded memory that is allocated to them. 
If the initial dataset is balanced, as it is assumed in existing incremental learning \cite{DBLP:conf/eccv/CastroMGSA18,DBLP:journals/corr/abs-1807-02802,DBLP:conf/cvpr/RebuffiKSL17}, the associated imbalance profile is binary, with new classes having a large number of images and past classes having a small but identical number of images.
If the dataset itself is imbalanced, as it is often the case in real contexts, the inherent imbalance is added to the incremental one and the resulting imbalance profile can be more complex.

Focus is put on calibration methods whose objective is to reduce the prediction bias between majority and minority classes.
We compare the following calibration methods: 
(1) isotonic regression \cite{DBLP:conf/kdd/ZadroznyE02} and Platt scaling \cite{platt1999probabilistic} which leverage initial scores to improve final predictions, 
(2) thresholding applied to the initial class probabilities in order to increase the predictions of rare classes \cite{DBLP:journals/nn/BudaMM18}, 
(3) nearest-exemplar-mean classifier \cite{DBLP:conf/cvpr/RebuffiKSL17} and balanced fine tuning \cite{DBLP:conf/eccv/CastroMGSA18} which were recently introduced as post-processing steps to reduce the effect of data imbalance in deep incremental learning and 
(4) two proposed methods which group classes in batches either as new vs. old or by image counts and then exploit the mean classification scores per batch for calibration.

While the focus is on methods which increase the performance on the test dataset, we also evaluate the intrinsic effect on model calibration. 
A model is said to be calibrated when the average confidence of the model is the same as its average accuracy.
Deep learning models have been shown to provide over-confident predictions that we do not match its accuracy\cite{DBLP:conf/icml/GuoPSW17}. 
In several scenarios, it is important to take into account the confidence of the model on the predictions. This is particularly the case, when decision has to be taken based on prediction of more than one model.  Well-calibrated models have the confidence levels aligned to the model accuracy and thus give valuable information of how likely the model is to be correct or incorrect.

The main findings are:
\begin{enumerate}
    \item the obtained results support the usefulness of a majority of post-processing methods for the reduction of bias toward majority classes
    \item when a bounded memory is available, the use of vanilla fine tuning followed by calibration is preferable to the widely used distillation loss \cite{DBLP:conf/eccv/CastroMGSA18,DBLP:journals/corr/abs-1807-02802,DBLP:conf/cvpr/RebuffiKSL17,DBLP:conf/cvpr/HouPLWL19,DBLP:conf/cvpr/WuCWYLGF19}.
    \item thresholding based calibration is most effective in providing overall improvement in accuracy, though it has detrimental affect on model calibration. The proposed methods provide consistent improvement in both model accuracy and model calibration. 
\end{enumerate}

We outline the structure of the remainder of the paper.
Section~\ref{sota} reviews relevant research from both incremental and imbalanced learning. 
Section~\ref{sec:problem} formalizes the incremental learning with imbalanced datasets problem. 
Section~\ref{sec:cal_methods} discusses calibration as an effective way to counter dataset imbalance and introduces the different calibration methods tested.
Section~\ref{sec:eval} compares the calibration methods to three strong incremental learning baselines and proposes an analysis of results in terms of accuracy (Subsection ~\ref{analysis-acc}) and the ability to provide calibrated predictions (Subsection ~\ref{analysis-ece}). Finally,
Section~\ref{conclu} presents the conclusions and perspectives related to the proposed contribution.

\section{Related works}
\label{sota}
Our research lies at the intersection of incremental and imbalanced learning.
Both topics witness a strong regain of interest after the advent of deep learning.
The two problems are well studied separately and the two issues are also simultaneously tackled for classical machine learning algorithms \cite{DBLP:journals/tsmc/PolikarUUH01,DBLP:conf/kdd/SyedLS99a}.
We analyze works from both areas and focus on challenges related to deep architecture complexity and to scalability. 
These properties are of utmost importance in applications such as visual content analysis.
The visual corpora to be analyzed evolve quickly and there is a need for updating the underlying classification models accordingly.

\subsection{Deep Incremental Learning}
\label{sota:il}
The main challenge in incremental learning is catastrophic forgetting \cite{mccloskey:catastrophic}, i.e. the tendency to forget previously learned information when new data is incorporated.
It occurs whenever access to old data is constrained or impossible.
If nothing is done to prevent this phenomenon, incremental learning predictions for past classes become random or nearly so. This is particularly the case of deep learning algorithms which heavily rely on labelled data. 
Recent incremental learning research exploits deep learning techniques for which is was shown that catastrophic forgetting is a major issue~\cite{bengio2013empirical}. 

Two main groups of methods have been proposed.
The first focuses on changing the neural net architectures to incorporate new knowledge.
Influential works include Growing a Brain \cite{DBLP:conf/cvpr/WangRH17}, progressive neural networks \cite{DBLP:journals/corr/RusuRDSKKPH16} or lifelong learning with a network of experts \cite{DBLP:conf/cvpr/AljundiCT17}. 
These methods are interesting but their complexity grows when new classes are added incrementally. 
Notably, inference time will become longer as the model grows and the scalability of these methods is consequently reduced. 

The second group adapts fine tuning to an incremental context by using a combination of classification and distillation losses \cite{DBLP:conf/eccv/CastroMGSA18,DBLP:journals/corr/abs-1807-02802,DBLP:conf/cvpr/RebuffiKSL17}.
These methods have constant model complexity, except for the classification layer which integrates new classes, and are more fitted for large scale content analysis.
Most of them require a bounded memory in order to partially avoid catastrophic forgetting.
The memory related constraint is more acceptable than model complexity growth when analyzing large datasets since the inference time is not influenced by the use of memory. 
Learning-without-Forgetting (LwF) is an influential method presented in  \cite{DBLP:conf/eccv/LiH16}.
The algorithm does not rely on past data and exploits knowledge distillation \cite{DBLP:journals/corr/HintonVD15} to reduce the discrepancy between activations of old classes from the original and new networks. 
iCaRL \cite{DBLP:conf/cvpr/RebuffiKSL17} builds on top of LwF in that it combines classification and distillation losses for each incremental state of the algorithm. 
A first important difference with LwF is that a bounded memory is allowed to store exemplars of old classes. 
As more classes are added, the number of images per old class is reduced to fulfill the memory constraint. Class exemplars are selected using a herding mechanism which gives priority to images that are closest to the class mean. 
A second difference is related to the classification mechanism. 
Instead of using the class activations of the deep models, a nearest-mean-of-exemplars is implemented.  
The $iCaRL$ average top-5 accuracy on $ILSVRC$ is 62.5\%.
An $iCaRL$ analysis \cite{DBLP:journals/corr/abs-1807-02802} indicates that the most important algorithm components are the bounded memory and the distillation loss. 
The herding mechanism and the nearest-mean-of-exemplars classification seem to matter less. 
Recently an end-to-end incremental learning scheme with a bounded memory in which predictions are provided by the deep model was introduced in \cite{DBLP:conf/eccv/CastroMGSA18}. 
The main modification compared to $iCaRL$ resides in the proposal of a loss function which includes separate distillation terms for each incremental batch.
In addition, data augmentation and balanced fine tuning used to reduce the effect of data imbalance between old and new classes. 
Top-5 accuracy on $ILSVRC$ is 69.4\%, to be compared with 62.5\% obtained by $iCaRL$. 
Interestingly, the use of herding to store exemplars is only marginally useful (0.5 points) compared to random selection.
This finding confirms the $iCaRL$ analysis conclusions from  \cite{DBLP:journals/corr/abs-1807-02802}. 
Existing methods reduce the effect of imbalance via the use of class exemplars \cite{DBLP:conf/cvpr/RebuffiKSL17} or balanced fine tuning \cite{DBLP:conf/eccv/CastroMGSA18}. We include these methods in our study of calibration methods.

A related approach $BiC$~\cite{DBLP:conf/cvpr/WuCWYLGF19} tackles the bias against old classes, by adding a linear layer with two learnable parameters after the classification layer. A small part of dataset is reserved to learn the parameters of the bias correction layer. The training is done in two steps, with the model and classifier weights are learnt first, before learning the bias correction layer using only the reserved dataset.  
In $LUCIR$~\cite{DBLP:conf/cvpr/HouPLWL19}, authors proposed three balancing constraints at the time of training to mitigate the imbalance bias between old and new classes. Firstly, they modify the distillation loss component using cosine normalization to counter larger weights and biases for new classes. Further, they exploit the observation that imbalance is less pronounced at the classifier margin to introduce a margin loss function which is less susceptible to imbalance. Finally, a less forget constraint is introduced which complements the distillation loss by encouraging the orientation of features extracted by current network to be similar to those by the original model.  
We include $BiC$~\cite{DBLP:conf/cvpr/WuCWYLGF19} and $LUCIR$~\cite{DBLP:conf/cvpr/HouPLWL19} as baselines to evaluate their performance on imbalanced datasets.

\subsection{Imbalanced Learning}
\label{sota:calib}

The methods that tackle imbalanced learning are based either on data sampling or classifier adaptation \cite{DBLP:journals/tkde/HeG09}.
Data-sampling methods mitigate the bias towards majority classes by balancing the training dataset. 
Balancing can be achieved either by undersampling the majority classes or by oversampling minority classes \cite{DBLP:journals/eswa/GuoLSMYB17}. 
The main risk of undersampling is that it leads to incomplete representation of classes. 
Informed undersampling \cite{DBLP:conf/icml/KubatM97} partially solves this problem by avoiding to select images which are close to class boundary. 
Oversampling methods are sensitive to overfitting. 
SMOTE  \cite{DBLP:journals/jair/ChawlaBHK02} is an influential solution to the problem and was improved in  \cite{DBLP:conf/pkdd/BellingerDJ16,DBLP:conf/icic/HanWM05,DBLP:conf/cidm/MaciejewskiS11}. It basically creates synthetic features for a dataset by applying simple arithmetic transformations to actual features. 
Undersampling and oversampling were recently studied in the context of deep imbalanced learning and were shown to have detrimental effect for imbalanced versions of the $ILSVRC$ dataset \cite{DBLP:journals/nn/BudaMM18}.
They will thus not be included among the calibration methods studied here.

Another line of research deals with imbalance at the classifier level. 
Thresholding (or post-scaling) modifies the decision threshold of the classifier to counter the bias toward majority classes. An interesting formulation of thresholding in  \cite{DBLP:journals/neco/RichardL91} where the outputs are modified using the prior class probabilities.
While very simple, thresholding outperforms a large array of data sampling and classifier level methods for object recognition using deep learning models in \cite{DBLP:journals/nn/BudaMM18}. 
It is thus a very competitive method and is evaluated here in the context of imbalanced incremental learning.
The authors of \cite{DBLP:conf/icml/GuoPSW17} show that the outputs of deep neural nets are miscalibrated, even for balanced datasets.
Miscalibration is likely to be even more important for imbalanced datasets. 
Two influential methods were shown to provide good results for calibration of classical machine learning algorithms \cite{DBLP:conf/icml/Niculescu-MizilC05}.
Isotonic regression \cite{DBLP:conf/kdd/ZadroznyE02} fits the raw classification results to a set of non-decreasing discrete set of values.
Platt scaling \cite{platt1999probabilistic} basically performs a logistic regression on the initial outputs of the classifier.
We use these two methods in our comparison of calibration methods.



\section{Problem formulation}
\label{sec:problem}

We consider $\mathcal{D}_N$ a labeled dataset $X_y,y_t \in \mathcal{X} \times \mathcal{Y}$ for $t=1,2,..T$ i.i.d realizations of random variables $\mathcal{X}, \mathcal{Y} \sim \mathbb{P}$, where $\mathbb{P}$ is the data distribution, $\mathcal{X}$ is the instance space and $\mathcal{Y}$ is the set of $N$ class labels $\{y_1,...,y_N\}$.
We denote $X_i =\{x_i^1,x_i^2, ..., x_i^{n_i}\}$ the set of $n_i$ instances for the class $y_i$ in the dataset $\mathcal{D}_N$. 
In a supervised classification problem, the objective is to learn a model $\mathcal{M} : \mathcal{X} \to \mathcal{Y} $ that maps an instance $x$ to a label vector $\hat{Y}$. By the following, we will denote $\hat{Y}$ as a set of the class prediction with $\hat{y_i}$ the prediction score for class $y_i$.

In an incremental learning setting, at each incremental state $k$, a set of $P_k$ new classes is added to the previous dataset with, for each new class $j$, a set of $n_j$ instances. The objective is thus to use $\mathcal{M}_{k-1}$, the model learned at the previous step, as input for an updated model $\mathcal{M}_{k}$ which classifies $N_k$ = $P_1 + P_2 + ... + P_{k}$ classes. Here, $N_k$ is the total number of classes that have been observed from the beginning. $\mathcal{M}_{k}$ is trained using a dataset $\mathcal{D}_{N_{k}}$ composed of all the instances of the $P_k$ new classes and only a restricted set of the instances of the $N_{k-1}$ old ones. In particular, we assume a bounded memory $B$ is available for the instances of the old classes in each incremental state. As a consequence, due to this limited memory size, $\mathcal{D}_{N_{k}}$ is by nature imbalanced and imbalance grows at each incremental state.

In this paper, we consider deep neural models $\mathcal{M}_k$ which include two main components. The first is a feature extractor $\mathcal{F}_k:\mathcal{X}_{N_k} \rightarrow \mathbb{R}^d$, with $d$ the size of the feature vector $f$. 
The second is a classifier $\mathcal{C}_k: \mathbb{R}^d \rightarrow \mathcal{Y}_{N_k}$ which outputs the classification scores $\hat{y_i}$ for the $N_k$ learned classes. The classification scores can then be converted to probability estimates $\hat{p_i}$ to ascertain the confidence of the model.
Depending on the calibration method used, $\mathcal{F}_k$ and $\mathcal{C}_k$ are either integrated in a single deep model or separated.

We are interested in learning incrementally over imbalanced datasets. 
In this context, the level of imbalance can be defined, for instance, by using a combination of mean ($\mu$) and standard deviation ($\sigma$) of the number of images per class. 
The higher the ratio between $\sigma$ and $\mu$ is, the stronger the imbalance of the dataset will be. 

\section{Calibration methods}
\label{sec:cal_methods}
Dealing with imbalance is important insofar the number of training samples per class often varies in real-life applications.
As a consequence, majority classes have better representations and are favored over minority ones. 
The application of calibration methods is an effective way to counter the effect of imbalance~\cite{DBLP:journals/nn/BudaMM18}.
Put simply, calibration attempts to boost predictions for minority classes in order to compensate for their weaker representation in the deep model.
We study different calibration methods proposed either in imbalanced or incremental learning literature.
Fine tuning algorithms for incremental learning update the model ${\mathcal{M}_{k-1}}$ at an incremental state $k$ with training examples from new classes $P_k$ and a bounded exemplar set from past classes $N_{k-1}$.
If the initial dataset is balanced, we assume that each class is represented by $S$ images.
The bounded memory thus generates a binary imbalance with old classes being represented by $\frac{B}{N_{k-1}}$ and new classes by $S$ images. 
We term this imbalance as incremental imbalance as it arises as a consequence of learning incrementally with a bounded memory.
In our context, a dataset imbalance due to the variable class image counts is added to the imbalance generated via incremental learning.
The imbalance profile is not binary anymore since new classes are represented by a variable number of images and the proposed calibration methods should take this into account.

\begin{figure}[t]
	\begin{center}
\includegraphics[width=0.7\textwidth]{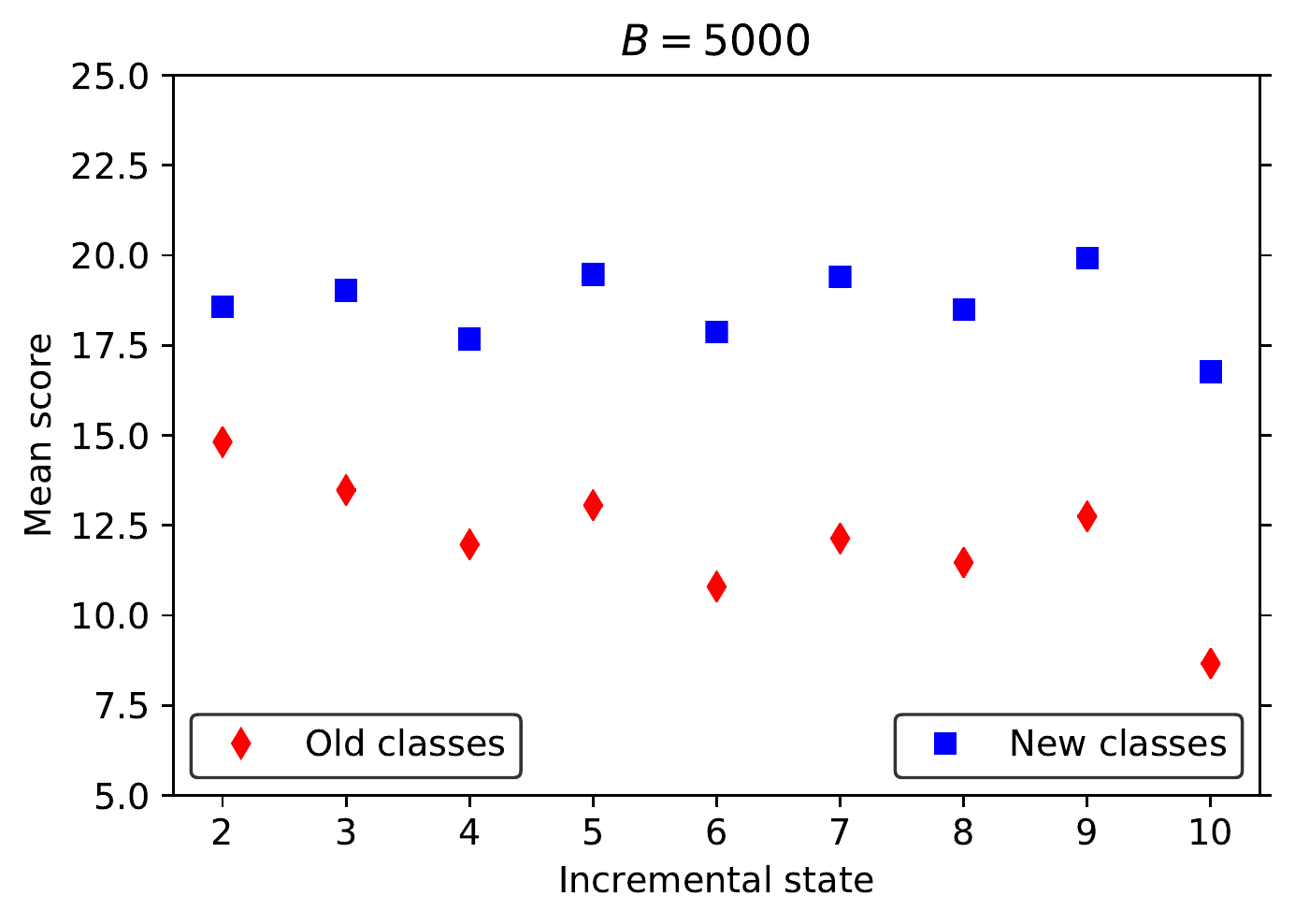}
	\end{center}
	\caption{Mean scores predicted by the model for samples from old and new classes using vanilla fine tuning for the $ILSVRC$ dataset. Training is done with 100 new classes added in each iteration using vanilla fine tuning, $B=5000$ bounded memory for old classes and $soft$ imbalance configuration as defined in Section~\ref{sec:eval}. The first, non-incremental, state is not represented.}
	\label{fig:means}
\end{figure}

When the model ${\mathcal{M}_{k}}$ is trained with a dataset affected by both incremental and dataset imbalance, it learns a feature extractor which is biased toward majority classes and performance is sub-optimal.
On average, the biased classifier associates higher scores to images from majority class than minority classes.
Figure~\ref{fig:means} illustrates this bias using the $ILSVRC$ dataset. 
The mean score of a class is computed using the predictions obtained for its samples from the training dataset. Then, we aggregate the average over the old and new classes to estimate the mean score of old and new classes.
We note that the mean scores of old classes are consistently lower than those of old classes for all incremental states.
Moreover, the difference tends to grow from left to right since the imbalance is higher in later incremental states.

We focus on fixed deep architectures and, in this case, the bias induced by imbalance needs to be reduced without increasing the complexity of the feature extractor $\mathcal{F}_k$ for the deep model $\mathcal{M}_k$.
\textbf{Consequently, calibration is defined as an adaptation of $\mathcal{C}_k$, the classification layer of $\mathcal{M}_k$, with the aim of reducing the bias toward majority classes.} 

We first present calibration methods which cover the main approaches from literature. 
Then we introduce two simple methods which leverage: (1) the prediction score means for old and new classes and (2) the distribution of the number of images per class.

\subsection{Isotonic regression calibration ($iso$)} 
Isotonic regression \cite{DBLP:conf/kdd/ZadroznyE02}  transforms the initial classifier predictions into a discrete set of calibrated scores. 
Since the number of available images per class reduces in incremental learning with bounded memory as more classes are added over time, isotonic regression exploits the training set  $\mathcal{D}_{N_k}$ .
The calibration is performed individually for each class and exploits the overlap between positive and negative examples for each class. 
$iso$ is based on the intuition that the distributions of predicted scores for positives and negatives examples are separable.
A discrete set of scores $R=\{0, ...,p_{l},..., 1\}$ is created where each discrete value represents a range of initial prediction scores. 
$p_l$ will be assigned to all initial predictions between two consecutive prediction boundaries $\hat{y_i}^{l-1}$ and $\hat{y_i}^{l}$ determined via isotonic regression, with $\hat{y_i}^{l-1} < \hat{y_i}^{l}$. 
The isotonic calibrated version of $\hat{y_i}$ is:
\begin{equation}
\hat{y_i}^{iso} = p_l,
\hspace{0.2cm} \textrm{if}\hspace{0.2cm} \hat{y_i}^{l-1} \leq \hat{y_i} < \hat{y_i}^{l}
\label{eq:is}
\end{equation}
Isotonic regression has the advantage of being non-parametric but requires a large number of positive samples per class to discretize probabilities in an efficient manner.
The method is applied as a post-processing step after $\mathcal{C}_k$.
The class predicted after calibration is the $argmax$ value of all calibrated class predictions given by Eq.~\ref{eq:is}.

\subsection{Platt calibration ($pl$)}
Platt scaling \cite{platt1999probabilistic} fits a logistic regression over the initial scores in order to reduce miscalibration.
The effect of $pl$ is effectuated at class level using one vs all selection of positive and negatives for each class. 
We write the calibrated score as: 
\begin{equation}
\hat{y_i}^{pl} = \frac{1}{1+\exp(A \hat{y_i} + C)}
\label{eq:ps}
\end{equation}
 where A and C are two parameters which need to be learned and $\hat{y_i}$ is the initial prediction of the $i^{th}$ class.
The incremental training set $\mathcal{D}_{N_k}$ is used to determine A and B by optimizing a max likelihood method. 
There is evidence that isotonic regression outperforms Platt scaling if enough examples per class are available \cite{DBLP:conf/icml/Niculescu-MizilC05}.
However, this finding was not tested for imbalanced datasets which may include a lot of minority classes, as it is the case here.
The class predicted after calibration is the $argmax$ value obtained by applying Eq.~\ref{eq:ps} to the initial scores predicted by $\mathcal{C}_k$.

\subsection{Thresholding based calibration ($th$)} 
Thresholding \cite{DBLP:journals/nn/BudaMM18} adjusts the prediction scores of a multi-class classifier by dividing the output of a class in $\mathcal{C}_k$ by its estimated prior probability. 
The calibrated score of is written as:
\begin{equation}
\hat{y_i}^{th} = \hat{y_i} \cdot \frac{\sum_{l=1}^{N_k} n_l}{n_i}
\label{eq:th}
\end{equation}
where $n_i$ is the number of images for the $i^{th}$ class and $\sum_{l=1}^{N_k} n_l$ is the total number of images in the training dataset $\mathcal{D}_{N_k}$.
By implementing Eq.~\ref{eq:th}, thresholding boosts the scores of minority classes. 
The smaller the number of samples of a class is, the stronger the boost it receives will be.
As we mentioned, a recent study of imbalanced learning for deep learning models showed that thresholding is highly efficient \cite{DBLP:journals/nn/BudaMM18}.
Its usefulness is theoretically supported by the fact that the outputs of a neural network correspond to Bayesian a posteriori probabilities \cite{DBLP:journals/neco/RichardL91}.
The class predicted after calibration is the $argmax$ value obtained over all predictions obtained with Eq.~\ref{eq:th}.

\subsection{Nearest-mean-of-exemplars calibration ($nem$)} 
The authors of $iCaRL$ \cite{DBLP:conf/cvpr/RebuffiKSL17} proposed nearest-mean-of-exemplars, an adaptation of nearest class-mean classifier \cite{DBLP:journals/pami/MensinkVPC13}, to counter the inherent imbalance in incremental learning.
The calibrated score of the $i^{th}$ class is written as:
\begin{equation}
\hat{y_i}^{nem} = \frac{1}{\lvert\lvert f(x) - \mu_i \lvert\rvert}
\label{eq:nem}
\end{equation}
where $f(x)$ is the $d$-dimensional feature of the test instance $x$ provided by penultimate layer of the incremental model $\mathcal{M}_k$; $\mu_i = \frac{1}{n_i}\sum_{l=1}^{n_i}f(x_l)$ - the mean feature of the exemplars available for the $i^{th}$ class.
Note that, in order to reduce the majority bias, $nem$ is performed after the selection of exemplars for new classes. 
Consequently, the number of exemplars per class is more balanced than if all samples were used and imbalance is reduced.
However, in the case of imbalanced datasets, there is no guarantee to achieve perfect balancing since a part of the classes might not have enough exemplars.
This is especially the case for a strong imbalance regime in which the number of samples per class varies strongly.
$nem$ calibration replaces the classification layer $\mathcal{C}_k$ of deep models by an external classifier which was explicitly designed to counter imbalance.
The class predicted for test instance $x$ after calibration is given by the $argmin$ function applied to the set of Euclidean distances computed for all classes using Eq.~\ref{eq:nem}.

\subsection{Balanced fine tuning calibration ($bal$)}
As an alternative to $iCaRL$ \cite{DBLP:conf/cvpr/RebuffiKSL17}, the authors of \cite{DBLP:conf/eccv/CastroMGSA18} propose an end-to-end incremental learning method.
The bias in favor of majority classes is reduced by introducing a second training step which trains all classes with the same number of exemplars. 
After the initial training which creates $\mathcal{M}_k$ using the imbalanced dataset $\mathcal{D}_{N_k}$, a model $\mathcal{M}_k^{bal}:\mathcal{X}_{N_k}^{bal} \rightarrow \mathcal{Y}_{N_k}$ is trained. 
$\mathcal{M}_k^{bal}$ exploits $\mathcal{D}_{N_k}^{bal}$ a balanced version of $\mathcal{D}_{N_k}$ which includes $\frac{B}{N_k}$ exemplar images for both old and new classes and is fine tuned starting from $\mathcal{M}_k$.
While this method is clearly appealing for balanced dataset, its usage in an imbalanced context is more challenging because there is no guarantee that there will be $\frac{B}{N_k}$ available for each class. 
We modify the approach slightly in that balanced fine tuning only learns the weights of the classification layer $\mathcal{C}_k^{bal}$, instead of fine tuning the entire model. 
This modification is done in order to make $bal$ calibration more comparable to the other calibration methods, which do not modify the feature extractor the deep model.
It is also motivated by the fact that initial experiments run with full fine tuning of $\mathcal{M}_k$ provided lower results than fine tuning the classification layer only.
Note that $bal$ has a higher computational cost at training time since it requires a supplementary training step.
The calibrated prediction of the $i^{th}$ class obtained with $bal$ can be written as:
\begin{equation}
   \hat{y_i}^{bal} = \mathcal{C}_k^{bal}(i)
   \label{eq:bal}
\end{equation}
where $\mathcal{C}_k^{bal}$ gives the output of classification layer of the balanced model $\mathcal{M}_k^{bal}$ for the $i^{th}$ class.
The class predicted after calibration is the $argmax$ value obtained over all classes using Eq.~\ref{eq:bal}.

\subsection{Batch mean based calibration ($mb$)}
\label{sub:mean}
The analysis of raw classification scores from Figure~\ref{fig:means} provides support for a bias in favor of new classes in imbalanced incremental learning. 
A simple way to reduce this imbalance is to exploit the mean prediction scores of new and old classes of incremental state $k$.
The calibrated score of the $i^{th}$ class is written as:
\begin{equation}
    \hat{y_i}^{mb} = \frac{\mu_{new}}{\mu_{old}} \hat{y_i}
    \label{eq:mb}
\end{equation}
where the means are defined as $\mu_{new} = \frac{1}{\sum_{l=1}^{P_k} n_l}\sum_{l=1}^{P_k}\sum_{q=1}^{n_l}\hat{y_q}$ 
and $\mu_{old} = \frac{1}{\sum_{l=1}^{N_{k-1}} n_l}\sum_{l=1}^{N_{k-1}}\sum_{q=1}^{n_l}\hat{y_q}$ for new and old classes respectively.
Note that here we hold out validation sets for new and old classes in order to compute their mean classification scores. 
Contrarily to $iso$ which works at class level, having a separate validation set is doable because $mb$ is applied at dataset level and the number of available samples is sufficient.
The class predicted after calibration is the $argmax$ prediction value obtained after applying Eq.~\ref{eq:mb}.

\subsection{Fisher-Jenks based calibration ($fj$)}
The mean based calibration operates at incremental batch level.
It disregards the fact that, due to dataset imbalance, some of the new classes might fall in the minority classes set. 
To counter this problem, we propose a calibration method which makes use of class image counts and of their associated classification score.
We use the Fisher-Jenks natural breaks method \cite{jenks77} to group classes.  
This method ensures an optimal distribution of a set of values in a predefined set of $L$ clusters.
It is thus appropriate to deal with the different imbalance profiles that occur in imbalanced incremental learning. 
In our case, the inputs given to Fisher-Jenks are the image counts $n_i$ associated to the $N_k$ classes learned in incremental state $k$. 
The calibrated score of the $i^{th}$ class is written as:
\begin{equation}
    \hat{y_i}^{fj} = \frac{\mu_{cl_L}}{\mu_{cl(i)}} \hat{y_i}
    \label{eq:fj}
\end{equation}
where $\mu_{cl(i)}$ is the mean prediction score of the Fisher-Jenks cluster which includes the $i^{th}$ class and $\mu_{cl_L}$ is the mean prediction score of the $L^{th}$ cluster with the largest number of instances per class.

$fj$ can be seen as a compromise between methods such as $iso$, which operate at class level, and $mb$, which works indifferently for all classes.
It groups classes depending on their sample distribution in order to have sufficient samples for a robust statistical distribution. 
Similar to the $mb$ method from Subsection~\ref{sub:mean}, the means are computed using a validation set. 
The number of Fisher-Jenks clusters is set using a cross-validation with the validation set. 
The class predicted after calibration is the $argmax$ prediction value obtained by applying Eq.~\ref{eq:fj} to all initial class predictions.

\section{Evaluation}
\label{sec:eval}
The experiments are designed to evaluate both kind of imbalances: dataset imbalance and incremental imbalance. 
All methods are evaluated with three large datasets designed for object, face and landmark recognition.
$Soft$ and $strong$ imbalance configurations are created to evaluate dataset imbalance.
Three bounded memory size are introduced for each dataset in order to test the robustness of calibration method with respect to this central parameter of incremental algorithms. 

\subsection{Baselines}
The calibration methods studied here are applied on top of a vanilla fine tuning backbone which is run iteratively for each incremental state in order to integrate new classes.
Naturally, vanilla fine tuning ($FT$ hereafter) is the main baseline used here. The selection of exemplars is based on the herding mechanism ~\cite{DBLP:journals/pami/MensinkVPC13}. 
To evaluate the usefulness of the proposed approach, we compare it to three competitive incremental learning methods:
\begin{itemize}
    \item $iCaRL$~\cite{DBLP:conf/cvpr/RebuffiKSL17} combines classification and distillation losses to counter catastrophic forgetting and uses a nearest exemplar mean classifier to counter imbalance between past and new classes.
    \item $BiC$~\cite{DBLP:conf/cvpr/WuCWYLGF19} introduces a linear layer at the end of the classification process to ensure fairness between past and new classes. A distillation term which is closer to the original formulation from~\cite{DBLP:journals/corr/HintonVD15} compared to $iCaRL$ is equally used.
    \item $LUCIR$~\cite{DBLP:conf/cvpr/HouPLWL19} proposes a combination of three elements to improve incremental learning. Cosine normalization is used for balancing the magnitudes of past and new class predictions. The distillation term is improved by handling feature vectors instead of raw scores. Finally, inter-class separation is favored in order to better separate embeddings of past and new classes.
\end{itemize}
\subsection{Datasets and methodology}
\label{sub:metho}
We evaluate the baselines and the calibration methods on the following datasets:
\begin{itemize}
\item $ILSVRC$  \cite{DBLP:journals/ijcv/RussakovskyDSKS15} is a subset of 1000 $ImageNet$ classes used in the\\
$ImageNet~LSVRC$ challenges. \\
\item $VGG Face2$ \cite{DBLP:conf/fgr/CaoSXPZ18} ($VGGF2$ below) focuses on face recognition. We select the 1000 classes with the largest number of associated images. \\ 
\item $Google Landmarks$ \cite{DBLP:conf/iccv/NohASWH17} ($LAND$ below) was built for landmark recognition and we again select 1000 classes with the largest number of associated images.

\begin{table}\centering
	\begin{tabular}{c|cc|cc|cc}
	\toprule
	
	& $\mu_{orig}$ & $\sigma_{orig}$ & $\mu_{soft}$ & $\sigma_{soft}$ & $\mu_{strong}$ & $\sigma_{strong}$ \\
    \midrule
    $ILSVRC$ & 1231 & 70 & 649 & 354 & 147 & 231 \\
    \hline
    $VGGF2$ & 492 & 49 & 266 & 129 & 97 & 120 \\
    \hline
    $LAND$ & 374 & 103 & 212 & 111 & 85 & 90 \\
    \bottomrule
	\end{tabular}
	\caption{Means and standard deviations of image counts  in the original datasets ($orig$) and the two imbalance configurations ($soft$ and $strong$).}
	\label{tab:stats}
\end{table}

\end{itemize}

The original amount of imbalance in these three datasets is weak, as shown in Table~\ref{tab:stats}. 
We introduce two imbalance configurations to evaluate behavior of the algorithms with different degrees of dataset imbalance:
\begin{itemize}
    \item $soft$ - randomly retains between 50 and the initial number of images for each class. 
    \item $strong$ - randomly retains between: 10 and 25 images for 300 classes, 26 and 75 for 300 classes, 76 and 100 for 200 classes and between 101 and the initial number of images for the remaining 200 classes. 
\end{itemize}
The corresponding means and standard deviations are reported in Table~\ref{tab:stats}. 
In the $soft$ configuration, slightly more than half of the original training data is kept and the standard deviation amounts to over 50\% of dataset means. 
With $strong$, we discard a wide majority of original data and the resulting imbalance is much stronger and the standard deviation becomes higher than the mean in each case. 

The evaluated calibration methods operate either at class level ($iso$, $pl$, $th$, $nem$, $bal$) or at an aggregate level which includes a subsets of the learned classes ($mb$, $fj$). 
For class level methods, we reuse the training images from the initial dataset as inputs for calibration.
This is necessary since the number of available images is reduced, especially for old and/or minority classes and most of the methods require a rather large amount of data to provide reliable results.
Consequently, the use of a validation subset would be suboptimal here.
When inputs from different classes are aggregated  in batches ($mb$ and $fj$), the use of a proper validation split becomes possible. 
We create validation sets using 10\% of the training data of old and new classes. 
We maintain the val/train split in the bounded memory $B$ to avoid mixing the training and validation exemplars in different incremental states. 
Note that the outputs of $C_{k}$ are used either in their raw form ($iso$, $pl$,$mb$, $fj$) or after transformation in probabilities by applying $softmax$ ($th$).
This choice is made in order obtain an optimal configuration of each algorithm.

The experimental setup is inspired by the one proposed in $iCaRL$ \cite{DBLP:conf/cvpr/RebuffiKSL17}. 
Each dataset of 1000 classes is split into $k=10$ incremental states. 
Each incremental state adds a batch of $P_k = 100$ classes to those that were already learned in states $1$ to $k-1$.
The same class ordering provided in $iCaRL$ \cite{DBLP:conf/cvpr/RebuffiKSL17} is reused for $ILSVRC$ and a random ordering of classes is created to form $VGGF2$ and $LAND$ states. 
The size of bounded memory $B$ was shown to have a central importance for the performance of incremental learning algorithms \cite{DBLP:conf/eccv/CastroMGSA18,DBLP:conf/cvpr/RebuffiKSL17}. 
To assess its influence on the proposed calibration methods, we report results with $B=\{5000, 10000, 20000\}$ exemplars stored in memory for each dataset and imbalance configuration.

A ResNet-18 architecture \cite{DBLP:conf/cvpr/HeZRS16} is used as a backbone for all experiments.  ResNet have been successful in allowing neural nets to be deeper by tackling the problem to stagnation of performance with addition of layers after some point. They employ residual mapping as the basis function which adds the input values to approximate the final function. ResNet-18 has one (7*7) and sixteen (3*3) convolutional layers in addition to two max pooling layers and a final linear classification layer. 
We used the publicly available $iCaRL$ TensorFlow implementation in  \cite{DBLP:conf/cvpr/RebuffiKSL17} with a binary cross-entropy loss and the original parameters proposed there. 
Vanilla fine tuning ($FT$) was implemented in Pytorch \cite{paszke2017automatic} using cross-entropy loss. 
The models were trained for 25 epochs with a initial learning rate of 0.1 at every incremental state and scheduled to decay by 0.1 when the loss plateaus out for 5 epochs.  
For $VGGF2$, face cropping is done with MTCNN \cite{DBLP:journals/spl/ZhangZLQ16} before further processing.
Training images are processed using randomly resized $224\times224$ crops and horizontal flipping and are normalized afterwards.

\subsection{Metrics}
\begin{itemize}
    \item Accuracy - the performance of different methods is evaluated using top-1 accuracy for each incremental step defined as:  
    \begin{equation}
        acc = 100 * \frac{1}{n} \sum_{i=1}^n argmax(\hat{Y)} == y_{i} 
    \end{equation}  
    , where $\hat{Y}$ is the set of  predicted score and $y_{i}$ is the true label for test sample $i$.    
    This measure is then averaged over all incremental states in order to obtain a single value for the entire incremental process.
    Note that averaged accuracy is the usual metric employed in incremental learning~\cite{DBLP:conf/cvpr/HouPLWL19,DBLP:conf/cvpr/RebuffiKSL17,DBLP:conf/cvpr/WuCWYLGF19}.
    The test dataset contains the same number of samples for each class. This gives equal importance to all the classes irrespective of class-distribution in the training dataset. The test sets include $50000$ images for $ILSVRC$ and $VGGF2$ and $20000$ for $LAND$. There are 50 images per class for the first two datasets and 20 for the latter.
    
    \item Expected Calibration Error (ECE)- is a metric to ascertain the difference between the model accuracy and confidence~\cite{DBLP:conf/icml/GuoPSW17}.  
    The estimation of accuracy and confidence is done by dividing the samples into bins based on confidence. In our implementation the number of bins $M$ are set to 20, to give 20 intervals of 1/M = 0.05 size from 0 to 1. $B_{m}$ are set of samples in the interval $m$ , with m = \{1,2 ... M\}  and $n$ is the number of samples in the test dataset.
    
    \begin{equation}
        ECE = \sum_{m=1}^{M} B_{m}/n * \lvert\lvert conf(B_{m}) - acc (B_{m})\lvert\lvert
        \label{eq:ece}
    \end{equation}
     \begin{equation}
            conf(B_{m}) = 1/B_{m} \sum_{i \in B_{m}} max(\hat{P}_{i})
    \end{equation}    
     \begin{equation}
            acc(B_{m}) = 1/B_{m} \sum_{i \in B_{m}} 1  (argmax(\hat{Y}_{i}) == y_{i} )
    \end{equation}     
    ,where $\hat{P}_{i}$ and $\hat{Y}_{i}$ are the set of predicted probability and score respectively and $y_{i}$ is the true label for test sample $i$.
The values of $ECE$ range from 0 to 1, with lower values indicating better model calibration.

\end{itemize}

\subsection{Analysis of Accuracy of Calibration methods}
\label{analysis-acc}

\begin{table*}[ht]
\begin{center}
\small\addtolength{\tabcolsep}{-2pt}
\begin{tabular}{|c|c|c|c|c|c|c|c|c|c|c|c|c|}
    \hline
$B$ & Dataset & \rotatebox{70}{$iCaRL$} & \rotatebox{70}{$LUCIR$} & \rotatebox{70}{$BIC$} & \rotatebox{70}{$FT$} & \rotatebox{70}{$FT_{is}$} & \rotatebox{70}{$FT_{pl}$} & \rotatebox{70}{$FT_{th}$} & \rotatebox{70}{$FT_{nem}$} & \rotatebox{70}{$FT_{bal}$} & \rotatebox{70}{$FT_{mb}$} & \rotatebox{70}{$FT_{fj}$} \\
\hline
    \multirow{3}{*}{5000}& $ILSVRC$ & 21.8 & \textbf{45.5} & 41.3 & 38.7 & 23.5 & 31.4 & 45.0 & 39.3 & 40.5 & 41.6 & 41.5\\
& $VGGF2$ & 61.0 & 84.4 & 78.7 & 81.4 & 42.7 & 65.5 & \textbf{85.4} & 81.9 & 81.4 & 84.9 & 85.1\\
& $LAND$ & 64.1 & 86.9 & 80.6 & 84.3 & 37.9 & 76.0 & \textbf{88.0} & 85.2 & 81.1 & 85.7 & 86.0\\
    \hline
    \multirow{3}{*}{10000}& $ILSVRC$ & 23.6 & 48.9 & 45.5 & 45.3 & 32.1 & 38.6 & \textbf{49.8} & 44.8 & 45.6 & 46.9 & 46.4\\
& $VGGF2$ & 62.1 & 86.9 & 80.3 & 86 & 66.1 & 76.5 & \textbf{88.0} & 85.2 & 82.2 & 87.4 & 87.7\\
& $LAND$ &65.7 & 88.9 & 82.1 & 88.9 & 53.9 & 84.6 & \textbf{90.7} & 88.2 & 85.8 & 89 & 89.2 \\
    \hline
    \multirow{3}{*}{20000}& $ILSVRC$ & 24.5 & 52.7 & 49.7 & 50.1 & 38.3 & 44.8 & \textbf{53.4} & 48.6 & 49.8 & 50.7 & 50.3 \\
& $VGGF2$ & 62.2 & 88.4 & 81.6 & 90.2 & 80 & 86.1 & \textbf{91.0} & 88.8 & 87.3 & 90.5 & 90.9 \\
& $LAND$ & 65.8 & 90.8 & 83.3 & 92.2 & 75.4 & 90.5 & \textbf{92.6} & 90.8 & 91.6 & 92.0 & 92.1 \\
    \hline

\end{tabular}
\end{center}
 \caption{Top-1 average accuracy for the $soft$ imbalance configuration and $B=\{5000, 10000, 20000\}$ bounded memory sizes. The first two columns represent $iCaRL$ \cite{DBLP:conf/cvpr/RebuffiKSL17} and vanilla fine tuning ($FT$), our baselines. The next two columns are calibrated versions of $FT$ as follows: $FT_{iso}$ - isotonic regression; $FT_{pl}$ - Platt scaling; $FT_{th}$ - thresholding; $FT_{nem}$ - nearest-mean-of-exemplars; $FT_{bal}$ - balanced fine tuning; $FT_{mb}$ - batch mean based calibration; $FT_{fj}$ - Fisher-Jenks based calibration. 
 Following \cite{DBLP:conf/eccv/CastroMGSA18}, accuracy scores are averaged over the incremental states of the system and the first, non-incremental, state is ignored.}
 \label{tab:soft}
\end{table*}

\begin{table*}[ht]
\begin{center}
\small\addtolength{\tabcolsep}{-2pt}
\begin{tabular}{|c|c|c|c|c|c|c|c|c|c|c|c|c|}
    \hline
$B$ & Dataset & \rotatebox{70}{$iCaRL$} & \rotatebox{70}{$LUCIR$} & \rotatebox{70}{$BIC$} & \rotatebox{70}{$FT$} & \rotatebox{70}{$FT_{is}$} & \rotatebox{70}{$FT_{pl}$} & \rotatebox{70}{$FT_{th}$} & \rotatebox{70}{$FT_{nem}$} & \rotatebox{70}{$FT_{bal}$} & \rotatebox{70}{$FT_{mb}$} & \rotatebox{70}{$FT_{fj}$} \\
\hline
    \multirow{3}{*}{5000}& $ILSVRC$ & 13.9 & 21.7 & 24.5 & 29.9 & 17.8 & 23.6 & \textbf{33.1} & 31.0 & 29.2 & 28.9 & 30.2  \\
& $VGGF2$ &  50.4 & 66.3 & 65.9 & 76.7 & 32.3 & 63.6 & \textbf{78.6} & 75.1 & 72.0 & 77.4 & 78.1\\
& $LAND$ & 57.5 & 77.2 & 73.3 & 80.8 & 36.6 & 74.7 & \textbf{82.4} & 80.7 & 80.4 & 80.7 & 81.3 \\
    \hline
    \multirow{3}{*}{10000}& $ILSVRC$ & 15.9 & 24.9 & 26.2 & 34.4 & 24.3 & 29.1 & \textbf{36.8} & 34.9 & 32.1 & 33.0 & 34.3 \\
& $VGGF2$ &  51.3 & 68.5 & 67.9 & 80.8 & 55.9 & 72.9 & \textbf{81.7} & 78.7 & 77.0 & 80.5 & 81.43 \\
& $LAND$ & 58.8 & 79.4 & 74.9 & 85.7 & 47 & 82.6 & \textbf{86.3} & 84.6 & 84.4 & 85.3 & 85.8 \\
    \hline
    \multirow{3}{*}{20000}& $ILSVRC$ & 16.2 & 27.0 & 27.1 & 37.5 & 29.1 & 33.0 & \textbf{39.4} & 37.6 & 34.7 & 36.2 & 37.3\\
& $VGGF2$ & 51.4 & 71.8 & 68.8 & 83.9 & 68.6 & 78.6 & \textbf{84.6} & 82.1 & 81.1 & 83.4 & 84.3 \\
& $LAND$ & 60.4 & 80.9 & 75.1 & 87.8 & 65.8 & 85.5 & \textbf{88.4} & 86.4 & 86.0 & 87.5 & 88.1 \\
    \hline  
\end{tabular}
\end{center}
 \caption{Top-1 average accuracy for the $strong$ imbalance configuration and $B=\{5000, 10000, 20000\}$ bounded memory sizes. See Table~\ref{tab:soft} for the description of the different methods presented.}
 \label{tab:strong}
\end{table*}

\begin{figure*}
\centering
\begin{minipage}{.499\textwidth}
  \centering 
  \includegraphics[width=.999\linewidth]{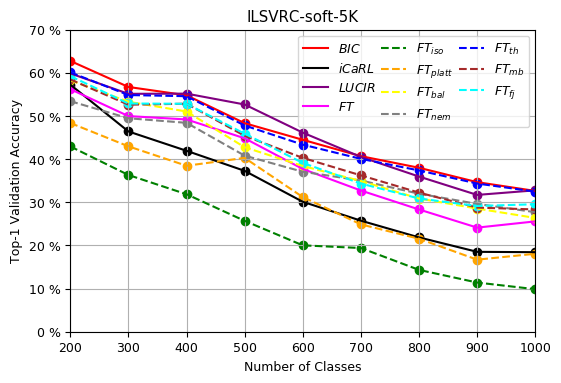}
\end{minipage}%
\begin{minipage}{.499\textwidth}
  \centering
  \includegraphics[width=.999\linewidth]{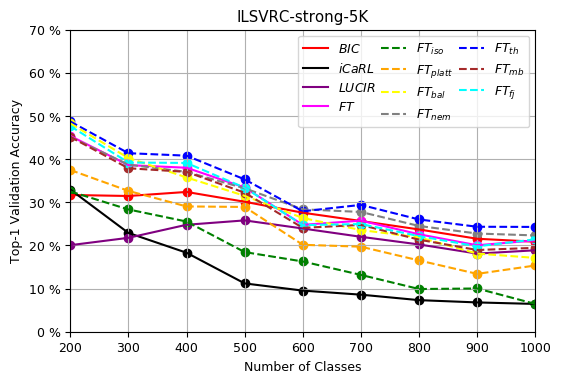}
\end{minipage}
	\caption{Top-1 accuracy for $ILSVRC$ with memory $B=5000$ in $soft$ (left) and $strong$ (right) imbalance configurations. To be aligned with the results from Tables~\ref{tab:soft} and~\ref{tab:strong}, only the incremental states are represented. (\textit{Best viewed in color.})
	}
	\label{fig:res}
\end{figure*}

\begin{table*}[ht]
\begin{center}
\small\addtolength{\tabcolsep}{-2pt}
\begin{tabular}{|c|c|c|c|c|c|c|c|c|c|}
    \hline
$B$ & Dataset & $FT$ & $FT_{is}$ & $FT_{pl}$ & $FT_{th}$ & $FT_{nem}$ & $FT_{bal}$ & $FT_{mb}$ & $FT_{fj}$ \\
\hline
    \multirow{3}{*}{5000}& $ILSVRC$ & \textbf{0.214} & 0.231 & 0.310 & 0.545 & 0.389 & 0.252 & 0.216 & 0.218  \\
& $VGGF2$ &  0.045 & 0.422 & 0.651 & 0.138 & 0.814 & 0.102 & \textbf{0.033} & 0.036  \\
& $LAND$ &  0.013 & 0.373 & 0.756 & 0.118 & 0.843 & 0.130 & \textbf{0.009} & 0.011\\
    \hline
    \multirow{3}{*}{10000}& $ILSVRC$ & \textbf{0.201} & 0.317 & 0.382 & 0.497 & 0.444 & 0.266 & 0.213 & 0.202\\
& $VGGF2$ & 0.022 & 0.803 & 0.857 & 0.087 & 0.883 & 0.089 & \textbf{0.021} & 0.022 \\
& $LAND$ & 0.011 & 0.533 & 0.841 & 0.091 & 0.874 & 0.111 & \textbf{0.009} & 0.012  \\
    \hline
    \multirow{3}{*}{20000}& $ILSVRC$ &  0.172 & 0.379 & 0.444 & 0.462 & 0.482 & 0.271 & 0.197 & \textbf{0.171}\\
& $VGGF2$ &  0.031 & 0.656 & 0.761 & 0.115 & 0.847 & 0.107 & \textbf{0.028} & 0.029 \\
& $LAND$ &  0.011 & 0.748 & 0.900 & 0.070 & 0.900 & 0.086 & \textbf{0.007} & 0.011 \\
    \hline    
\end{tabular}
\end{center}
 \caption{Expected Calibration Error for the $soft$ imbalance configuration and $B=\{5000, 10000, 20000\}$ bounded memory sizes. The first columns represent vanilla fine tuning ($FT$), our baseline. The next columns are calibrated versions of $FT$ as follows: $FT_{iso}$ - isotonic regression; $FT_{pl}$ - Platt scaling; $FT_{th}$ - thresholding; $FT_{nem}$ - nearest-mean-of-exemplars; $FT_{bal}$ - balanced fine tuning; $FT_{mb}$ - batch mean based calibration; $FT_{fj}$ - Fisher-Jenks based calibration. ECE are averaged over the incremental states of the system and the first non-incremental state is ignored.}
 \label{tab:ece-soft}
\end{table*}

\begin{table*}[ht]
\begin{center}
\small\addtolength{\tabcolsep}{-2pt}
\begin{tabular}{|c|c|c|c|c|c|c|c|c|c|}
    \hline
$B$ & Dataset & $FT$ & $FT_{is}$ & $FT_{pl}$ & $FT_{th}$ & $FT_{nem}$ & $FT_{bal}$ & $FT_{mb}$ & $FT_{fj}$ \\
\hline
    \multirow{3}{*}{5000}& $ILSVRC$ & 0.276 & 0.239 & 0.288 & 0.627 & 0.345 & 0.426 & 0.339 & \textbf{0.275}  \\
& $VGGF2$ & \textbf{0.060} & 0.318 & 0.632 & 0.204 & 0.746 & 0.197 & 0.063 & 0.068  \\
& $LAND$ &  0.018 & 0.360 & 0.743 & 0.165 & 0.799 & 0.179 & \textbf{0.017} & 0.018\\
    \hline
    \multirow{3}{*}{10000}& $ILSVRC$ & \textbf{0.286} & 0.174 & 0.233 & 0.662 & 0.307 & 0.405 & 0.337 & 0.288\\
& $VGGF2$ & \textbf{0.044} & 0.554 & 0.725 & 0.175 & 0.782 & 0.174 & 0.057 & 0.054 \\
& $LAND$ &  0.015 & 0.465 & 0.822 & 0.129 & 0.838 & 0.157 & \textbf{0.012} & 0.018 \\
    \hline
    \multirow{3}{*}{20000}& $ILSVRC$ &  0.263 & 0.287 & 0.326 & 0.601 & 0.372 & 0.453 & 0.324 & \textbf{0.260} \\
& $VGGF2$ & \textbf{0.044} & 0.681 & 0.783 & 0.148 & 0.816 & 0.157 & 0.055 & 0.051 \\
& $LAND$ &  0.014 & 0.653 & 0.850 & 0.110 & 0.856 & 0.127 & \textbf{0.010} & 0.015 \\
    \hline    
\end{tabular}
\end{center}
 \caption{Expected Calibration Error for the $strong$ imbalance configuration and $B=\{5000, 10000, 20000\}$ bounded memory sizes.}
 \label{tab:ece-strong}
\end{table*}

The obtained results are presented in Tables \ref{tab:soft} and \ref{tab:strong}. 
A detailed view of top-1 accuracy for the incremental states of $ILSVRC$ with $B=5000$ bounded memory for $soft$ and $strong$ imbalance configurations is provided in Figure~\ref{fig:res}.

The performance level of the presented methods is much lower than that of non-incremental and balanced learning. 
We trained a ResNet-18 non- incrementally and using the full $ILSVRC$ dataset and obtained a top-1 accuracy of $73.0\%$.
The non-calibrated accuracy ($FT)$ obtained for $ILSVRC$ with memory $B=20000$ are $50.1\%$ and $37.5\%$ for $soft$ and $strong$ imbalance configurations. 
The best results obtained for the same settings after calibration are $53.4\%$ and $39.4\%$ respectively. 
If the allowed memory is $B=5000$, performance goes from $38.7\%$ and $29.9\%$ (non-calibrated $FT$) to $45.0\%$ and $33.1\%$ ($FT_{th}$) for $soft$ and $strong$ configurations respectively.

Intuitively, performance for $soft$ imbalance (Table~\ref{tab:soft}) is higher compared to that for $strong$ imbalance (Table~\ref{tab:strong}). 
The difference between the two configurations is largest for $ILSVRC$, the most difficult dataset among the three tested. 
With a memory of $B=10000$ exemplars, the difference in performance between $soft$ and $strong$ configurations for $FT$ is $10.9\%$, $5.2\%$ and $3.2\%$ for $ILSVRC$, $VGGF2$ and $LAND$ respectively. 
The size of the memory has also a strong influence on results. 
For instance, the performance of $FT$ on $ILSVRC$ for bounded memories $B=\{5000, 10000, 20000\}$ reaches $38.7\%$, $45.3\%$ and $50.1\%$ in the $soft$ imbalance configuration.

The combined effect of incremental learning and dataset imbalance is thus strong and, while calibration is useful, the problem remains an open one. 
The difference between $soft$ and $strong$ imbalance configurations is also well illustrated in Figure~\ref{fig:res}. 
These detailed results show that the induced imbalance has a particularly important effect in early incremental states. 
This is normal since the importance of dataset imbalance is reduced in later incremental states, where the incremental imbalance due to the bounded memory $B$ acts upon a large majority of classes. 

The analysis of individual calibration methods shows that isotonic regression ($FT_{iso}$) and Platt calibration ($FT_{pl}$) have detrimental effect for both $soft$ and $strong$ imbalance configurations. 
Both methods rely heavily on the number of available class samples. 
The negative influence of $iso$ and $pl$ is larger for lower memory sizes and, within each $B$ size, for later incremental states.
This is probably an effect of lack of sufficient data in order to obtain a stable parametrization of the methods. 
The behavior of $iso$ and $pl$ in imbalanced incremental learning is different from the one previously reported in \cite{DBLP:conf/icml/Niculescu-MizilC05}.
There are two main differences between the two studies: (1) the algorithms used are different (deep models here and shallow models in \cite{DBLP:conf/icml/Niculescu-MizilC05}) and (2) the amount of data available for calibration which is much smaller here.  

Thresholding ($FT_{th}$) improves performance for all tested configurations. 
This method has the largest positive effect among all methods tested in a wide majority of cases.
$th$ performs score post-processing and is less dependent of the number of samples than $iso$ and $pl$. 
It provides the largest improvements for $B=5000$, the memory setting which corresponds to the largest imbalance for the three visual tasks with $soft$ and $strong$ configurations.
The results obtained for $th$ confirm those presented in \cite{DBLP:journals/nn/BudaMM18} for imbalanced learning.
They indicate that this simple calibration method should be considered in priority for the implementation of imbalanced incremental learning applications.

Nearest-mean-of-exemplars ($FT_{nem}$) has contrasted performance. The method is beneficial for the object recognition task ($ILSVRC$), although with lower effect for $B=20000$.
For face recognition task ($VGGF2$) and landmarks ($LAND$) gain is observed only for $soft$ imbalance at 5000 budget .
For $ILSVRC$, $nem$ is more useful for the $strong$ imbalance configuration than for the $soft$ one.
The method works on top of the penultimate layer of the deep model but is highly dependent of the number of samples available to compute the individual class means, a property shared with $iso$ and $pl$.
Note also that $FT_{nem}$ is equivalent to an $iCaRL$ version in which the distillation loss was ablated.
The authors of $iCaRL$ \cite{DBLP:conf/cvpr/RebuffiKSL17} report that $nem$ classification  has positive influence over a direct use of deep model predictions in all configurations tested in their paper.
A main difference is that those tests were done with classification and distillation losses, with larger memory and with datasets that are initially balanced.
The effect of $nem$ is more contrasted for the imbalanced datasets tested here with a vanilla fine tuning backbone with exemplars selected based on moving mean.


Balanced fine tuning ($FT_{bal}$) has a negative effect in most configurations.
The methods provides improvement over $FT$ for $ILSVRC$ at 5000 and 10000 budgets with $soft$ imbalance. The effect is particularly detrimental for $strong$ imbalance, 
Note that the reported $bal$ performance is obtained by fine tuning only the classification layer of the incremental deep models $\mathcal{M}_k$.
Balanced fine tuning ($FT_{bal}$) is performed by creating a balanced dataset, leading to much smaller datasets, especially if there is more dataset imbalance in addition to incremental imbalance. This would explain the sub-optimal performance as the dataset imbalance is increased.

Batch mean based calibration ($FT_{mb}$) improves performance over $FT$ for all settings with $soft$ imbalance, while being comparative to $FT$ for $strong$ imbalance. 
As for $th$ , the gains are larger for lower memory size and for $ILSVRC$, the hardest visual task tested here. 
$mb$ and $fj$ have comparable results for $soft$ imbalance configurations, while $fj$ gives slightly better results for $strong$ imbalance.
$mb$ is the simplest of all calibration methods tested since it only exploits mean predictions for old and new classes. 
It only accounts for the incremental imbalance as it groups new and old classes together, regardless of their image counts.

Fisher-Jenks based calibration ($FT_{fj}$) is a refined version of $mb$ in which both the incremental and dataset imbalance are taken into account when clustering classes. 
The advantage of such clustering is more obvious for $strong$ imbalance configurations, where the dataset imbalance is more important compared to $soft$ imbalance. 
Its performance is better than that of $bal$ and $nem$ for both $soft$ and $strong$ imbalance.
$ft$ globally has lower performance than $th$ calibration.

A statistical analysis of the $LUCIR$ and $FT_{th}$ reveal that $FT_{th}$ is significantly better for $strong$ imbalance regime as compared to $soft$ imbalance. We compute the p-values over the accuracies at each incremental batch to ascertain the significance in the incremental setting. For $ILSVRC$ dataset, the p-value at budgets 5000, 10000 and 20000 are 0.91 , 0.84 and 0.85 for $soft$ imbalance as compared to 0.02 , 0.006 and 0.005 for $strong$ imbalance. Similarly for $Land$ dataset the p-value for $soft$ imbalance is at 0.43, 0.10 and 0.001 as compared to 0.001, 0.003 and 0.0028 for $strong$ regime. For $VGGF2$, the p-values for $soft$ imbalance is at 0.119, 0.038 and 0.019 as compared to 0.0041, 0.0042 and 0.0017 for $strong$ regime.

A final interesting observation is that $iCaRL$ performance lags well behind that of $FT$ for all datasets and tested configurations. Further, $FT$ baseline is competitive with $LUCIR$ and $BIC$, at $soft$ imbalance, while being clearly the preferable option in the $strong$ imbalance regime.
This comparison is contrary to the conclusions of \cite{DBLP:conf/cvpr/RebuffiKSL17}, where $FT$ has significantly worse performance compared to $iCaRL$.
However, that evaluation was biased insofar $iCaRL$ was using a memory of past classes while $FT$ results obtained in absence of this memory.
Our results indicate that, when running a fair comparison, the simpler $FT$ method is clearly a better suited backbone for incremental learning with bounded memory than the state-of-the-art backbone which combines classification and distillation losses \cite{DBLP:conf/eccv/CastroMGSA18,DBLP:journals/corr/abs-1807-02802,DBLP:conf/cvpr/RebuffiKSL17}.

\subsection{Analysis of Expected Calibration Error}
\label{analysis-ece}
The results for Expected Calibration Error $ECE$ for the calibration methods are presented in Tables \ref{tab:ece-soft} and \ref{tab:ece-strong}. 
The first main observation is that the value of $ECE$ for $FT$ is higher for $ILSVRC$ dataset as compared to the other two datasets. This is explained by the fact that $VGGF2$ and $LAND$ are easier to learn and $FT$ provides significantly higher accuracy for these two datasets. Hence, the confidence is matched with high performance, which is not the case for $ILSVRC$.

Isotonic regression ($FT_{iso}$) and Platt calibration ($FT_{pl}$) provide very high $ECE$ values as compared to $FT$, particularly for $VGGF2$ and $LAND$ datasets. A look at $LAND$ at $soft$ imbalance with 10000 budget, shows the accuracy at  $53.9\%$ and $84.6\%$ for $FT_{iso}$ and $FT_{pl}$, whereas the $ECE$ is $0.533$ and $0.841$ respectively.
This allows us to infer that the confidence of probabilities after $FT_{iso}$ and $FT_{pl}$ calibration is quite low, and the models actually under-calibrated. This can be partly explained by limited number of positives instances for a class, and high number of negative instances in one-vs-all calibration used in Isotonic Regression and Platt Scaling. 

The results for $FT_{nem}$ are similar to $FT_{iso}$ and $FT_{pl}$ with very high values of $ECE$, particularly when the accuracy is high. We draw similar conclusions that the model is under-calibrated for $FT_{nem}$ as well. Note that for $FT_{nem}$, the scores are calculated as the inverse of the distance to the class mean in the feature space, which are then used to derive the probabilities using the softmax function. The accuracy for $FT_{bal}$ are slightly lower than $FT$, and this is also reflected in $ECE$ values of $FT_{bal}$ which are slightly higher than $FT$. 

$FT_{th}$ provides the most improvement in accuracy out of all the calibration method. $th$ mitigate the bias towards minority classes by calibrating the score for a class depending on the number of samples in the given class. It provides better performance by increasing the confidence of minority classes, though it also makes the model more mis-calibrated. $ECE$ score for $FT_{th}$ is consistently higher than its $FT$ counterpart. This shows that $FT_{th}$ provides improvement in overall accuracy, but at the some expense of calibration of model. 

The proposed methods $FT_{mb}$ and $FT_{fj}$ provides the best calibration out of all the calibration methods. The calibration is quite similar to $FT$ with ECE values being quite close to ones for for $FT$. This is an interesting results since $FT_{mb}$ and $FT_{fj}$ are the only methods which provide improvement in overall accuracy while not adversely affecting the calibration of the model.

\section{Conclusion}
\label{conclu}
We performed a study of score calibration methods in an incremental and imbalanced deep learning setting which was not explored before.
Calibration methods selected from both imbalanced and incremental learning streams of research were thoroughly compared using three visual tasks, two imbalance configurations and three bounded memory sizes for incremental learning.
The obtained results indicate that, while calibration is certainly useful, imbalanced class incremental learning remains an open problem. 
They also show that both dataset imbalance and memory size have an important impact on performance.
This is particularly true for object recognition, the most difficult of the three tested tasks, and for lower memory sizes.
 
The performance of the evaluated calibration methods is variable. Isotonic regression and Platt calibration, which were shown to work well when enough data per class is available \cite{DBLP:conf/icml/Niculescu-MizilC05}, have a negative effect on results here.
This behavior is explained by the scarcity of available data when working in an incremental setting. 
Nearest-mean-of exemplars \cite{DBLP:conf/cvpr/RebuffiKSL17} and balanced fine tuning \cite{DBLP:conf/eccv/CastroMGSA18}, the calibration methods introduced in recent incremental learning works, have contrasted and negative effects respectively.
Note that, after initial experiments, an adaptation of balanced fine tuning was performed so as to fine tune only the classification layer instead of the entire network as done in \cite{DBLP:conf/eccv/CastroMGSA18}.
The batch mean based and Fisher-Jenks calibration methods introduced here have a positive effect in most of the configurations. Fisher-Jenks behaves slightly better than mean based calibration. This is explained by the fact that the first method models both dataset and incremental imbalance while the second models only incremental imbalance.
The best performance in terms of accuracy is obtained by thresholding based calibration, which uses the prior class probabilities to augment the scores of minority classes. 
An analysis of model calibration after the calibration method shows that overall $FT_{mb}$ and $FT_{fj}$ provide the best model calibration, while also tacking the imbalance.

Finally, the results also show that vanilla fine tuning is a better backbone for class incremental learning with bounded memory compared to a fine tuning which exploits both classification and distillation losses.
The performance gap between the two approaches is significant and we advocate that future developments in class incremental learning should use vanilla fine tuning as baseline.

The reported results are interesting and can be developed along the following lines: 
(1) improve the vanilla fine tuning backbone using recent results in imbalanced learning \cite{DBLP:journals/nn/BudaMM18}; 
(2) explore other score calibration methods and
(3) test the calibration methods using different model architectures

\bibliographystyle{unsrt}  
\bibliography{references}

\end{document}